# Performance Analysis of Metaheuristic Algorithms for Quadratic Assignment Problem

Reza Tavakkoli-Moghaddam; Zohreh Raziei

School of Industrial Engineering, College of Engineering

University of Tehran

Tehran, Tehran 50122, Iran

Siavash Tabrizian

Department of Industrial Engineering

Sharif University of Technology

Tehran, Tehran 50122, Iran

## Abstract

The quadratic assignment problem (QAP) is a combinatorial optimization problem. This problem belongs to the class of NP-hard problems. So, it is difficult to solve in the polynomial time even for small instances. Research on the QAP has thus focused on obtaining a method to overcome this problem. The heuristics and metaheuristics algorithm are prevalent solution methods for this problem. This paper is one of comparison study to apply different metaheuristic algorithms for solving the QAP. One of the most popular approaches for categorizing metaheuristic algorithms is based on search strategy that included (1) Local search improvement metaheuristics. (2) Global search based metaheuristics. The matter that distinguishes this paper from the other is the comparative performance of local and global search (both EA and SI) metaheuristic algorithms that consist of Genetic Algorithm (GA), Particle Swarm Optimization (PSO), Hybrid GA-PSO, Grey Wolf Optimization (GWO), Harmony Search Algorithm (HAS) and Simulated Annealing (SA). Also, one improvement heuristic algorithm (2-Opt) is used to compare with others. The PSO, GWO and 2-Opt are improved to achieve the better comparison toward the other algorithms for evaluation. In order to analysis the comparative advantage of these algorithms, eight different factors are presented. By taking into account all these factors, the test is implemented in 6 test problems of the Quadratic Assignment Problem Library (QAPLIB) from different sizes. Another contribution of this paper is measuring strong convergence condition for each algorithm in a new way.

## Keywords

## 1. Introduction

The quadratic assignment problem (QAP) is a combinatorial optimization problem, which is presented by (Koopmans et al., 1957). The general QAP locates facilities with respect to the cost minimization of the placing facility and distances from other facilities where flow exists between every pair of facilities.

The QAP belongs to the class of NP-hard problems. Therefore, exact solutions have been incompatible for QAP in large size instances, because they need a large amount of computational time for solving this problem (Bayat & Sedghi, 2009).

(Loiola et al., 2007) proposed a survey about tendencies for 50 years studies of QAP. They categorize this study in the application, theory and algorithms. This tendency is shown in the period 1990 to 2005 in Figure 1. This paper has shown the number of publications in algorithm design is more than two others subjects. Also, recent surveys on QAP proposed by (Bhati & Rasool, 2014) that described some application of QAP which have been applied to real world problems. (Bayat & Sedghi, 2009) presented a complete survey on QAP about the variance formulation of the problem and different solution methods. Heuristic and meta-heuristic algorithms are the best guide for obtaining the feasible or a good solution for large instances, but this solution usually is near optimal.

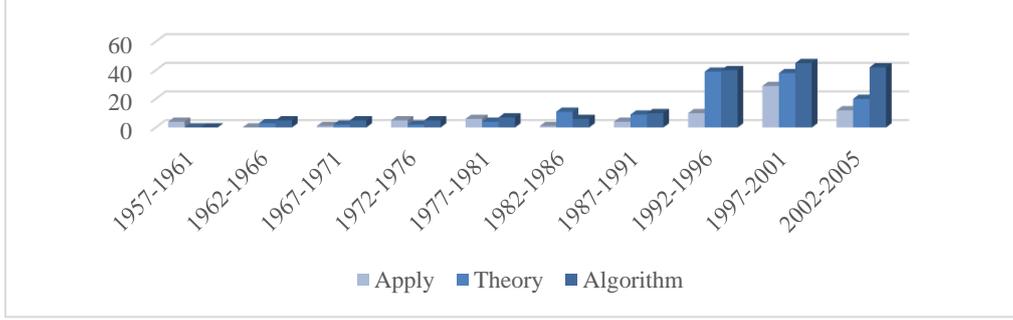

**Figure 1: Number of 59 year publication in QAP for this categorizes (Loiola et al., 2007)**

There have been used several meta-heuristic to solve QAP that some examples are mentioned like: Scatter Search (SS) (Cung et al., 1997), Greedy Randomized Adaptive Search Procedure (GRASP) (Oliveira et al., 2004). Also, some hybrid meta-heuristic algorithm are capable of obtaining the best solution. (Gambardella et al., 1997) solved QAP with hybrid ACO algorithm and simple local search called (HAS-QAP). They were compared this algorithm with TS, reactive TS, hybrid GA and SA. Their comparisons show the hybrid GA has better performance to solve this problem. (Lim et al., 2000) proposed a hybrid genetic algorithm with a deterministic local search procedure to solve QAP. (Tseng & Liang, 2006) proposed a hybrid meta-heuristic algorithm that integrated ACO, GA and local search method and called ANGEL method. They showed the ability of ANGEL for obtaining optimal solutions with a 90 % success rate.

As an example of new studies, (Tasgetiren et al., 2013) proposed three meta-heuristic to solve QAP. They used Iterated Greedy Algorithm (IGA), Discrete Differential Evolution (DDE) algorithm and migrating birds' optimization (MBO) algorithm by using the test problem of QAPLIB (Burkard et al., 1997). The results show better performance of IGA algorithm. (Kaviani et al., 2014) presented a hybrid meta-heuristic based on TS and SA to solve QAP by using QAPLIB's test problems.

To the best of our knowledge, this paper is one of comparison study for analysis the performance of different meta-heuristics for the QAP. The matter that distinguishes this paper from the other is the comparative performance of local and global search (both EA and SI) metaheuristic algorithms that consist of Genetic Algorithm (GA) (Holland, 1975), Particle Swarm Optimization (PSO) (Kennedy & Eberhat, 1997), Hybrid GA-PSO, Grey Wolf Optimizer (GWO) (Mirjalili et al., 2014), Harmony Search Algorithm (HAS) (Geem et al., 2001) and Simulated Annealing (SA) (Hwang, 1988). Also, one improvement heuristic algorithm (2-Opt) is used to compare with others. In addition, the PSO, GWO and 2-Opt are improved to achieve the better comparison toward the other algorithms for evaluation with new metrics. In the Section 2, a brief description about QAP and QAPLIB is presented. The description about algorithms are presented in Section 3. Section 4 is shown the applied metrics for evaluation algorithms. In Section 5 the numerical results and analysis are provided and the conclusion and future research are provided in Section 6.

## 2. Quadratic Assignment Problem (QAP) formulation

The QAP formulation first presented by (Koopmans & Beckmann, 1975).

$$min \ z = \sum_{i,j=I} \sum_{p,k=1} f_{ij} \, d_{kp} x_{ik} x_{jp} \qquad (1)$$

s.t.

$$\sum_{i \in n} x_{ij} = 1 \qquad 1 \leq j \leq n \qquad (2)$$

$$\sum_{j \in n} x_{ij} = 1 \qquad 1 \leq i \leq n \qquad (3)$$

$$x_{ij} \in \{0,1\} \qquad 1 \leq i,j \leq n \qquad (4)$$

Where $F = [f_{ij}]_{n \times n}$ is the matrix of flow between facility $i$ and $j$, $D = [d_{kl}]_{n \times n}$ is the matrix of distance between location $k$ and $l$. By considering the cost of allocation of facilities to locations, the problem can be formulated as follows:

$$min \ z = \sum_{i,j=l} \sum_{p,k=1} f_{ij} d_{kp} x_{ik} x_{jp} + \sum_{i,k \in n} b_{ik} x_{ij} \quad (5)$$

s.t.

$$(2)\text{-}(4) \quad (6)$$

Where $B = [b_{ik}]_{n \times n}$ the matrix of cost of allocation facility $k$ to location $i$ and $c_{ijkl}$ is the cost obtained from $f_{ij}.d_{kl}$.

### 2.1 Quadratic Assignment Problem Library (QAPLIB)

The QAPLIB was first published in 1991. It is the collection of test bed for QAP from different subscriber. According to the website information, it's getting started from Graz University of Technology then preserved by the University of Pennsylvania. This library consists of 137 test problems from 15 subscriber source. These test problems cover the real-world as a random test with a size range from 10 to 256. Because of the continuing demand for these test problems a major update was provided by (Burkard et al., 1994). The size of all test problems selected for this paper are between 15 and 150 from Scr (M. Scriabin & R.C. Vergin), Wil (M.R. Wilhelm & T.L. Ward) and Tho (U.W. Thonemann & A. Bölte) test problems.

### 3. Heuristic and Metaheuristic Algorithms

In this study, heuristic and metaheuristic approaches are used to solve QAP. Heuristic algorithms are divided into three classes: (1) Construction algorithm, (2) Improvement algorithm, (3) Hybrid algorithm (Heragu, 2008). In the class of improvement algorithm, the Local Search Heuristic (LSH) based on the 2-Opt algorithm is used in this study. One of the most popular approaches for categorizing metaheuristic algorithms is based on search strategy that included (1) Local search improvement metaheuristics. (2) Global search based metaheuristics. The global search strategy's metaheuristics are population based that included Evolutionary Algorithm (EA) and Swarm Intelligence (SI) (Blum & Roli, 2003). According to this categorizing, the metaheuristic algorithms that used in this paper included: Genetic Algorithm (GA), Particle Swarm Optimization (PSO), Hybrid GA-PSO, Grey Wolf Optimizer (GWO), Harmony Search Algorithm (HAS) and Simulated Annealing (SA). Also, one improvement local search heuristic algorithm based on 2-Opt is utilized in order to compare with others. The PSO, GWO and 2-Opt are improved to achieve the better comparison toward the other algorithms for evaluation. So, the mechanism of these algorithms is shown in the following sub-sections.

#### 3.1 Local search heuristic (LSH)

A 2-Opt algorithm is a simple improvement algorithm. Also, it is used as a simple local search algorithm. The 2-Opt and 3-Opt algorithm first used to solve the Traveling Salesman Problem (TSP) (Deo & Kowalik, 2006). In this paper, the local search heuristic (LHS) consists of a 2-Opt algorithm with Inversion Mutation (IVM) operator (Fogel, 1990) of genetic algorithm to improve the diversification in 2-Opt. The procedure of this algorithm for QAP is shown in Figure 2.

#### 3.2 Metaheuristic Algorithm

In this subsection, the metaheuristic algorithms are explained that have different features from the original version. The motivation for Genetic algorithm is the mechanism of natural selection and natural genetics as first articulated by (Holland, 1975). GA consists of a family of parallel, randomized-search optimization heuristics. According to the rich literature review of GA, it refers to the reproduction, mutation, crossover, and selection mechanism. In this study, the selected mechanism consists of the meta-ordering crossover operator (MOX) (Asveren & Molitor, 1996) as crossover operator. For mutation mechanism, Insertion Mutation (ISM) (Fogel 1988), Inversion Mutation (IVM) (Fogel 1990) and Exchange Mutation (EM) (Bankhaf 1990) are used. Also, there are many mechanisms for selection individual solutions from the population. The roulette wheel selection is chosen to use in proposed GA.

| | |
|---|---|
| $S$ = allocation, $Z$=fitness function<br>**begin** *(Alorithm)*<br>1. *Initialize.* Set $Z^* = Z$, $S^* = S$<br>2.  **while** *(t<Max iteration)*<br>3.    *Set i=1 and j=i+1*<br>4.    *Exchange i and j*<br>5.    *Calculated Z'*<br>6.    **if** *(Z' < Z)*<br>7.       $Z^* = Z'$, $S^* = S'$<br>8.    **end** *(if)*<br>9.    **if** *(j<num facility)*<br>10.       *j=j+1*<br>11.    **end** *(if)*<br>12.    **if** *(j=num facility & i<num_facility)*<br>13.       *i=i+1*<br>14.    **end** *(if)*<br>15.    *Go to 3*<br>16.    $S$ = *Inversion mutation* $(S^*)$<br>17.    **if** *( i=num facility)*<br>18.       *Go to 20*<br>19.    **end** *(if)*<br>20.    **if** *( $S^* \neq S$)*<br>21.       $Z^* = Z$, $S^* = S$<br>22.    **Other**<br>23.       *Go to 1*<br>24.    **end** *(if)*<br>25. **end** *(while)*<br>**end** *(Algorithm)* | **begin** *(Alorithm)*<br>1. **for** *each particle $i \in 1,...,I$* **do**<br>2.    *Randomly initialize $x_i$*<br>3.    *Randomly intialize $v_i$ (or just set $v_i$ equal to zero)*<br>4.    *Set $p_i=x_i$*<br>5. **end**<br>6. **Repeat**<br>7. **for** *each particle $i \in 1,...,I$* **do**<br>8.    *Evaluate the fitness of particle i by $f(x_i)$*<br>9.    *Update $p_i$*<br>10.    $p_i' $ = *perfom mutation on $p_i$*<br>11.    **if** *( $f(p_i') < f(p_i)$)*<br>12.       $p_i \leftarrow p_i'$<br>13.    **end** *(if)*<br>14.    *Update $\hat{p}$*<br>15.    $\hat{p}' $ = *perfom mutation on $\hat{p}$*<br>16.    **if** *( $f(\hat{p}') < f(\hat{p})$)*<br>17.       $\hat{p} \leftarrow \hat{p}'$<br>18.    **end** *(if)*<br>19.    **for** *each dimension j* **do**<br>20.       *Update apply velocity*<br>21.    **end** *(for)*<br>22.    *Update apply position*<br>23.    *Update coefficient $c_1$ and $c_2$*<br>24. **end** *(for)*<br>25.**Until** *some convergence criteria is satisfied*<br>**end** *(Algorithm)* |
| **Figure 2:  The pseudo-code of the Local Search Heuristic** | **Figure 3: The pseudo-code of the Hybrid GA-PSO Algorithm** |

PSO developed by (Kennedy & Eberhat, 1997) as a parallel evolutionary computation technique. In order to improve the convergence rate of this algorithm (also Hybrid GA-PSO), the coefficient $(\overrightarrow{c_1})$ and $(\overrightarrow{c_2})$ are updated (decreased) at each iteration that are fixed in each iteration of original PSO.

In general, it has a high probability that PSO falls into the trap of local optimal solution in various iterations. In order to improve the solutions of PSO, the mutation operators of GA are added for obtaining the partial best solution and global best solution. The procedure of the hybrid GA-PSO algorithm shown in Figure 3.

Harmony Search is a music-inspired meta-heuristic algorithm and it was developed by (Geem et al., 2001) the aim of this algorithm is to search for a perfect state of harmony in the musician's improvisation process to find the optimal solution in the optimization process.

GWO is a population based algorithm, which is inspired by grey wolves (Canis Iupus) (Mirjalili et al., 2014). This algorithm is provided based on hierarchy, tracking, encircling, and attacking mimics of prey. According to the hierarchical feature of GWO, the fitness solution considers as the alpha (α). Also, the second and third best solution is named beta (β) and delta (δ) and other candidate solutions are omega (ω). The hunting or optimization provides by α, β and δ wolves. The ω wolves are the follower. In order to obtain the new position of wolves, two random parameters $\overrightarrow{r_1}, \overrightarrow{r_2} \in U(0,1)$ that used in encircling prey vector are produced for each type of wolves (α, β and δ) separately (it has been refused to show the pseudo-code of improvement GWO algorithm, because of the limited number of pages). This mechanism causes the better convergence of best, mean and worst objective function of this algorithm.

## 4. Evaluation Metrics

The metaheuristic algorithms utilize different strategies for searching in solution area, and their random operators, it is essential to analyze their performances for different problems. This paper is one of comparative studies that apply different metaheuristic algorithms for solving the QAP. In order to analyze the comparative advantage of these algorithms, firstly all the parameters of algorithms should be tuned, then because the stochastic behavior of these algorithms, these algorithms are run several times. There are several indicators to evaluate the common feature of each algorithm. A number of indicators include, rate of efficiency, robustness of computing, rate of convergence, the deviation of solution and etc. In this paper, eight factors are presented that consist of the mean of best, average, and worst cost; variance of best, average, and worst cost; run time; and the rate of efficiency. By taking into account all these factors, the test is implemented in 6 test problems of the Quadratic Assignment Problem Library (QAPLIB) from different sizes (Burkardet al., 1997).

In an analysis of the rate of efficiency, the time of each iteration is measured. It is better to measure the time between iteration $t$ and $t+1$ ($\lambda_{s+1}$), because, the complexity of the first iteration has a lot of tolerance from the complexity of lasts iteration. In order to compare the rate of efficiency of each algorithm together, choose one of the three formulations that shown below:

| | |
|---|---|
| $\lambda_{min} = min_s\{\lambda_{s+1}\}$ | Minimum number of time\iteration to find new solutions. |
| $\bar{\lambda} = (\sum_{s=1}^{S-1} \lambda_{s+1})/S$ | Mean number of time\iteration to find new solutions. |
| $\lambda_{max} = max_s\{\lambda_{s+1}\}$ | Maximum number of time\iteration to find new solutions. |

The algorithm is more robust if the variance of $\lambda_{s+1}$ is less than other algorithms. In order to compare the robustness of several algorithms, it is better to use Goodness of Fit Test. Each algorithm with closest value of $\lambda_{t+1}$ to uniform distribution between ($\lambda_{min}, \lambda_{max}$) is selected as robustness algorithm.

Another important factor for comparison the performance of metaheuristic algorithms is strong convergence condition. One of the contribution of this paper is measuring this factor for each algorithm in a different way. It is measured by calculating the difference of the maximum variation coefficient of $n$ iteration of the algorithms and the minimum of that, and then when this value less than $\delta$ for $k$ times (the value of $n$, $\delta$, and $k$ are tuned) the algorithm will achieve to the strong convergence. The procedure is shown in Figure 4.

```
if (i>n)
    Max coefficient of varition = Max (√Variance best (i-n to i)) / Mean (Mean (i-n to n))
    Min coefficient of varition = Min (√Variance best (i-n to i)) / Mean (Mean (i-n to n))
    Gap coefficient of varition = Max coefficient of varition − Min coefficient of varition
    if (Gap coefficient of varition<δ )
        k ← k+1
    end (if)
end (if)
```

**Figure 4: Procedure for evaluation strong convergence**

## 5. Experimental Results

In this section, first, the result of eight factors and best objective function are illustrated for seven algorithms on the six test problems from QAPLIB.

**Table 1: Comparison of results obtained of seven algorithms**

| Q.M | T.P | GA | PSO | GA-PSO | GWO | HS | SA | LSH |
|---|---|---|---|---|---|---|---|---|
| **Best Obi.** | | 27953 | 32465 | 28523 | 38639 | 35024 | 25570 | 27964 |
| **M.B** | Scr15 | 28333 | 32651 | 30019 | 38459 | 35201 | 26436 | 27964 |
| **M.A** | | 28694 | 41902 | 47110 | 38447 | 51506 | 29308 | 29612.6 |
| **M.W** | | 28358 | 61080 | 61782 | 38506 | 649448 | 32829 | 34470.1 |
| **V.B** |  | 1302400 | 6380300 | 8321700 | 7504200 | 122560 | 2180852 | 0 |
| **V.A** |  | 5618100 | 1794100 | 1257800 | 7465600 | 12.256 | 6907100 | 26966.6 |
| **V.W** |  | 8622400 | 3259000 | 25915000 | 8207200 | 0 | 13708727 | 499990.6 |
| **Efficiency** |  | 0.097 | 0.037 | 0.091 | 0.019 | 0.001 | 0.211 | 0.02 |
| **Time (s)** |  | 19.306 | 7.385 | 17.691 | 2.372 | 0.165 | 42.025 | 0.6 |
| **Best Obj.** |  | 56805 | 72608 | 69603 | 87727 | 83250 | 55484 | 26983 |
| **M.B** |  | 61265 | 74879 | 70702 | 97393 | 83250 | 57562 | 26983 |
| **M.A** |  | 62270 | 104150 | 105870 | 98573 | 112280 | 64658 | 28973.6 |
| **M.W** |  | 62663 | 131160 | 131620 | 99276 | 135861 | 73757 | 33514.7 |
| **V.B** | *Scr20* | 3264200 | 39505000 | 33286000 | 39109000 | 0 | 17706383 | 0 |
| **V.A** |  | 5742000 | 68891000 | 58640000 | 50077000 | ~ 0 | 51433653 | 37165.8 |
| **V.W** |  | 7236400 | 104150000 | 105510000 | 51316000 | 0 | 73091688 | 313168.1 |
| **Efficiency** |  | 0.077 | 0.039 | 0.086 | 0.029 | 0.001 | 0.226 | 0.01 |
| **Time** |  | 15.3263 | 7.828 | 17.262 | 5.888 | 0.236 | 36.397 | 0.6 |
| **Best Obj.** |  | 127145 | 152350 | 137110 | 155071 | 157474 | 123697 | 126413 |
| **M.B** |  | 133820 | 149210 | 140040 | 157870 | 158010 | 129609 | 126413 |
| **M.A** |  | 134460 | 169710 | 168400 | 167090 | 171150 | 134772 | 132168.56 |
| **M.W** |  | 134660 | 182150 | 180520 | 170960 | 180037 | 140072 | 151503 |
| **V.B** | *Tho40* | 3798900 | 67598000 | 65227000 | 65274000 | 827240 | 45448292 | 0 |
| **V.A** |  | 5272600 | 72830000 | 71529000 | 74517000 | 82.724 | 45448292 | ~ 0 |
| **V.W** |  | 5858000 | 88150000 | 86891000 | 74511000 | 0 | 87270864 | 0 |
| **Efficiency** |  | 0.1259 | 0.044 | 0.115 | 0.038 | 0.006 | 0.254 | 0.1 |
| **Time** |  | 50.378 | 17.276 | 45.946 | 15.317 | 2.331 | 50.561 | 6.6 |
| **Best Obj.** |  | 24859 | 26353 | 25817 | 27165 | 4803239 | 24808 | 24744 |
| **M.B** |  | 25273 | 26515 | 26207 | 27170 | 4803400 | 25205 | 24744.2 |
| **M.A** |  | 25303 | 27620 | 27628 | 27397 | 4901200 | 25568 | 25080.9 |
| **M.W** |  | 25313 | 28306 | 28278 | 27502 | 5001074 | 25897 | 25661.1 |
| **V.B** | *Wil50* | 180960 | 1435400 | 1473900 | 1511900 | 77887 | 21092 | 20.8 |
| **V.A** |  | 227600 | 1533600 | 1533500 | 1483400 | 778871 | 28349 | 301.7 |
| **V.W** |  | 247350 | 1622100 | 1618900 | 1531700 | 0 | 330025 | 4526.9 |
| **Efficiency** |  | 0.126 | 0.051 | 0.109 | 0.045 | 0.002 | 0.309 | 0.1 |
| **Time** |  | 63.152 | 25.386 | 54.206 | 22.508 | 7.545 | 61.558 | 9.9 |
| **Best Obj.** |  | 138636 | 146260 | 142557 | 147795 | 147492 | 140088 | 137830 |
| **M.B** |  | 140150 | 146310 | 144120 | 147920 | 147650 | 142067 | 137831.5 |
| **M.A** |  | 140200 | 149560 | 149590 | 148840 | 149830 | 142949 | 138173.3 |
| **M.W** |  | 140220 | 151310 | 151270 | 149150 | 151177 | 143837 | 138681.9 |
| **V.B** | *Wil100* | 3350300 | 19619000 | 20394000 | 19944000 | 26785 | 3780856 | 1879.6 |
| **V.A** |  | 3759500 | 20379000 | 20384000 | 20217000 | 2.679 | 4631953 | 13049.7 |
| **V.W** |  | 3906700 | 20941000 | 20923000 | 20294000 | 0 | 5134321 | 79048.4 |
| **Efficiency** |  | 0.171 | 0.094 | 0.215 | 0.086 | 0.003 | 0.748 | 0.1 |
| **Time** |  | 187.341 | 103.286 | 236.3168 | 94.567 | 3.666 | 148.791 | 97.4 |
| **Best Obj.** |  | 4187055 | 4713723 | 4514603 | 4803499 | 4803239 | 4360028 | 4197255 |
| **M.B** |  | 4289266 | 4719800 | 4581600 | 4803900 | 4803400 | 4494691 | 4197416.8 |
| **M.A** |  | 4292088 | 4885000 | 4891000 | 4864300 | 4901200 | 4544923 | 4203400.7 |
| **M.W** | *Tho150* | 4292955 | 4971800 | 4975500 | 4887500 | 5001047 | 4598950 | 4210389.9 |
| **V.B** |  | $1.5 \times 10^{13}$ | 16366000 | $1.94 \times 10^{12}$ | $1.65 \times 10^{12}$ | 778870 | $1.39 \times 10^{10}$ | 18145132.2 |
| **V.A** |  | $1.6 \times 10^{13}$ | $1.70 \times 10^{12}$ | $1.71 \times 10^{12}$ | $1.71 \times 10^{12}$ | 778871 | $1.51 \times 10^{10}$ | 73767720.9 |
| **V.W** |  | $1.7 \times 10^{13}$ | $1.71 \times 10^{12}$ | $1.79 \times 10^{12}$ | $1.72 \times 10^{12}$ | 0 | $1.68 \times 10^{10}$ | 293261029 |

| | | | | | | | |
|---|---|---|---|---|---|---|---|
| **Efficiency** | 0.295 | 0.165 | 0.389 | 0.155 | 0.005 | 1.533 | 0.267 |
| **Time** | 413.437 | 230.465 | 544.847 | 216.894 | 7.545 | 305.059 | 373.082 |

**QM:** Quality Measurement; **MB:** Mean of Best cost; **MA:** Mean of Average cost; **MW:** Mean of Worst cost; **VB:** Variance of Best cost; **VA:** Variance of Average cost; **VW:** Variance of Worst cost; **TP:** Test Problem

In addition to the results that mentioned above, the first most important diagrams are convergence diagrams of best, worst, and average. These diagrams show the decline rate of the variation of the solutions of the algorithms through iterations. The second important factor is the time at which the three diagrams are met (it has been refused to show these diagrams, because of the limited number of pages).

According to the results in Table 1, the best objective function of SA is better than the others for small and medium size, but the convergence of the objective function of the GA is better through the time. However, the efficiency shows the dispersion in last iterations for both GA and SA. The dispersion in CPU time of the iterations shows that in last iterations the both algorithms unstable. In this case, take into account the mean of best, average and worst variances and efficiency diagram LSH has the best behavior among these algorithms. One of the best local search algorithm for QAP is 2-OPT and inserting the mutation operator in this algorithm is improved this algorithm in an excellent fashion. The efficiency rate for the LHS is so good and it shows the significant stability of the algorithm. After LSH, the efficiency rate is shown the stability of GA, too. The stability is important for algorithms, because the stability is a factor that shows the intelligence of the algorithm. The stable algorithms show the equivalent results during every use of the algorithm.

The run time of PSO is lower than others because the decline rate of the PSO algorithm through the iterations is high. However, the best and mean of best, average and worst objective function is increased. Since the population of the algorithm is not shown the equivalent behavior through the time the algorithm is trapped in local optimums easily, and also the algorithm bind in stagnation in the feasible area.

By adding the mutation operators when obtaining the partial best solution and global best solution in PSO, the best solution is improved.

The HS has good results for efficiency and runtime, but one of the weaknesses in HS algorithm is disability of the improvement mechanism that is used in GWO, the convergence of its objective function is better than the HS, but the rate of efficiency (or dispersion in its efficiency diagram) show the disability of the algorithm in preserve its stability through the iterations is greater than HS.

The results are shown each algorithm has special characteristics. It can be seen that GA algorithm is the best algorithm with respect to the best mean value of the objective function. However, the SA has the best objective function for small and medium size. Also, by taking into account the rate of efficiency and the runtime, LSH has the best behavior among these algorithms. With respect to the convergence rate GA algorithm shows the better behavior among the presented algorithms.

The results of strong convergence condition for several runs of algorithms (with Tho150) are shown in Table 2. In the several run of algorithm, The SA has a better objective function, but took much run time and number of iterations. Also, by comparing the diagram of strong convergence of best, worst, and average and the diagram of efficiency of SA (Figure 15 and 16) with the diagrams of GA (Figure 5 and 6), GA-PSO (Figure 9 and 10) and LSH (Figure 17 and 18), It can be observed that the strong convergence of GA for best, mean and worst objective function is better than the other algorithms. But, the diagram of efficiency shows the dispersion in last iterations for LSH is less than the others.

**Table 2: The results strong convergence**

| S.C | Best Obj. | | | Num. of iteration | | | Run time | | |
|---|---|---|---|---|---|---|---|---|---|
| | *Max* | *Mean* | *Min* | *Max* | *Mean* | *Min* | *Max* | *Mean* | *Min* |
| GA | 4175060 | 4167537.5 | 4151565 | 4375 | 3883.5 | 2950 | 606.0399 | 533.3539 | 409.0175 |
| PSO | 4709245 | 4678861 | 4641674 | 1047 | 909.25 | 749 | 72.2529 | 63.0459 | 52.2574 |
| GA-PSO | 4172771 | 4162922 | 4155533 | 5484 | 5059.67 | 4532 | 751.8737 | 684.9693 | 604.2 |
| GWO | 4792310 | 4777367 | 4740151 | 387 | 367.75 | 345 | 28.02158 | 25.8379 | 23.6841 |
| HS | 4807364 | 4803414 | 4793364 | 551 | 327.25 | 202 | 4.4668 | 2.8946 | 1.5650 |
| SA | 4123061 | 4120449.8 | 4116408 | 12743 | 11886 | 10584 | 3731.9213 | 3451.4616 | 3069.2456 |
| LSH | 4177303 | 4169327 | 4159696 | 46 | 45.5 | 45 | 35.37876 | 32.52807 | 30.89023 |

**S.C:** Strong Convergence.

In comparison of Hybrid GA-PSO with PSO, the results are shown the best objective function for Hybrid algorithm is improved, but the PSO is better strong convergence for objective function. Also, the dispersion in iterations of PSO for efficiency diagram is less than the Hybrid algorithm.

In the comparison of the HS and GWO, HS has good results for efficiency and runtime, but one of the weaknesses in HS algorithm is disability of the algorithm in meliorate of the worst value and average value of the algorithm.

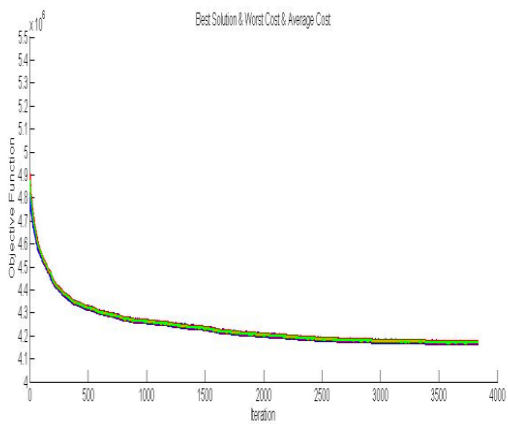

**Figure 5: Strong converges rate of Best, average and worst solution for GA**

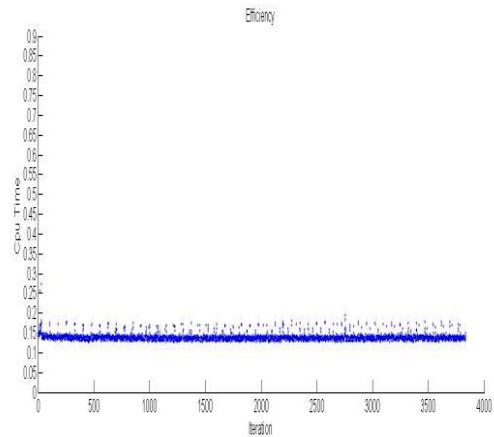

**Figure 6: The time efficiency of strong convergence for GA**

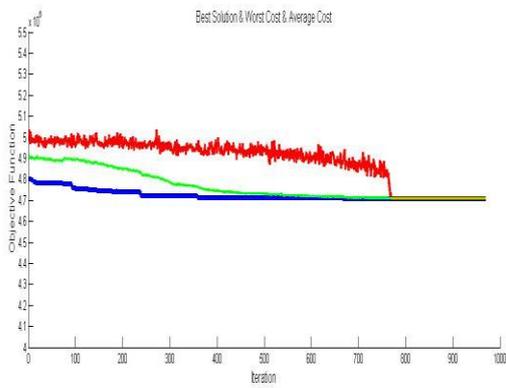

**Figure 7: Strong converges rate of Best, average and worst solution for PSO**

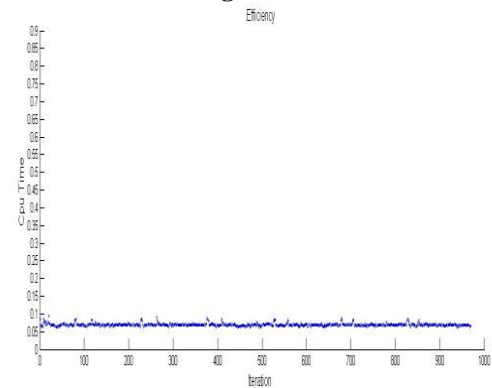

**Figure 8: The time efficiency of strong convergence for PSO**

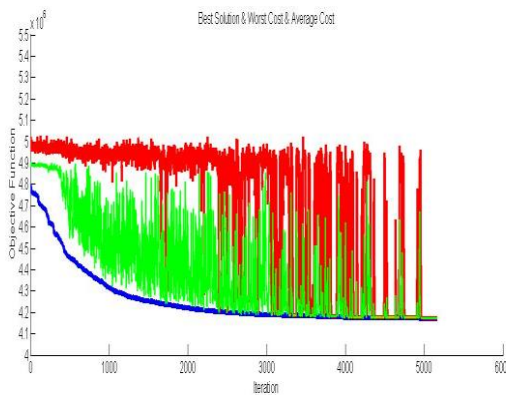

**Figure 9: Strong converges rate of Best, average and worst solution for Hybrid GA-PSO**

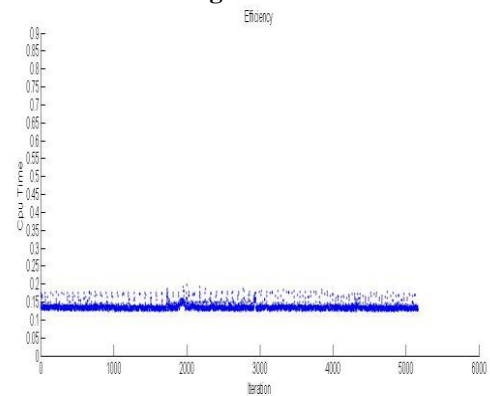

**Figure 10: The time efficiency of strong convergence for Hybrid GA-PSO**

With the improvement mechanism that is used in GWO, it can be seen the strong convergence of the objective function for GWO better that HS. But, that the GWO the dispersion in its efficiency diagram of the GWO is shown the disability of the algorithm in preserving its stability through the iterations.

In order to show the trend of strong convergence of all algorithms, the results variance of objective function for 50 initial iteration is shown in Figure 19. Each algorithm that has low variance fluctuation, will converge faster. The convergence rate of GA from the first iteration is appropriate. Also, GWO and LHS show the good convergence after giving improvement mechanisms. According to the previous result of SA, this algorithm need to long run time for reaching the strong convergence. The PSO convergence is not occurring in the initial iterations of this algorithm. It can achieve strong convergence after a lot of iterations with effect of improvement mechanism that mentioned above.

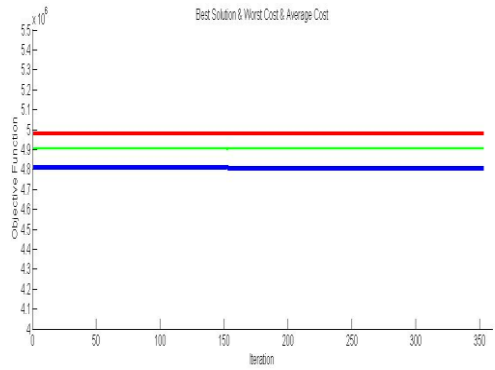

Figure 11: Strong converges rate of Best, average and worst solution for HS

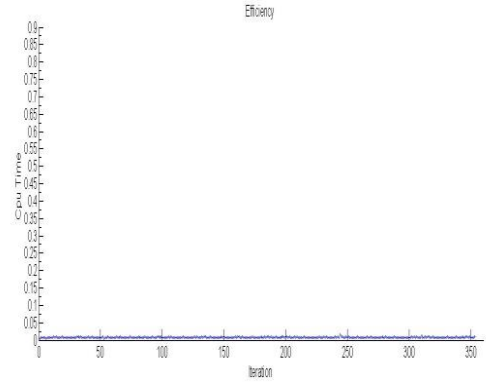

Figure 12: The time efficiency of strong convergence for HS

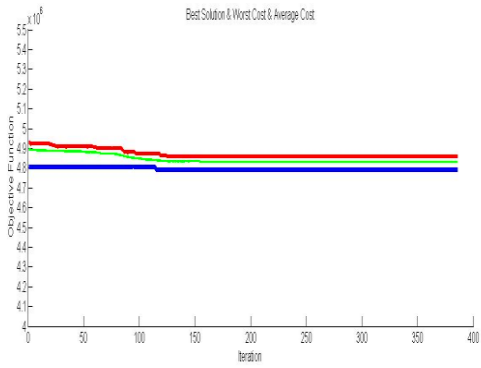

Figure 13: Strong converges rate of Best, average and worst solution for GWO

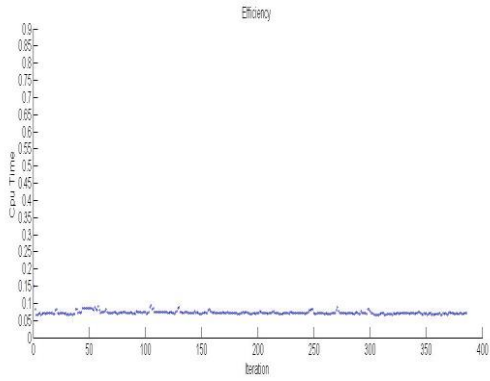

Figure 14: The time efficiency of strong convergence for GWO

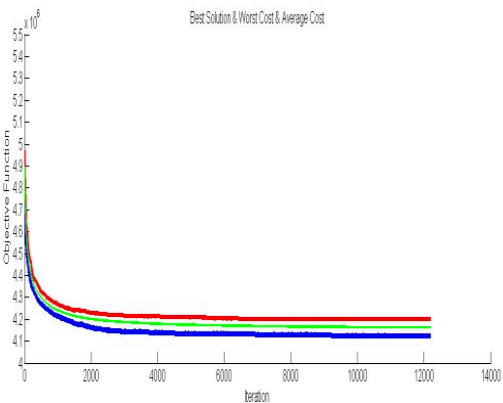

Figure 15: Strong converges rate of Best, average and worst solution for SA

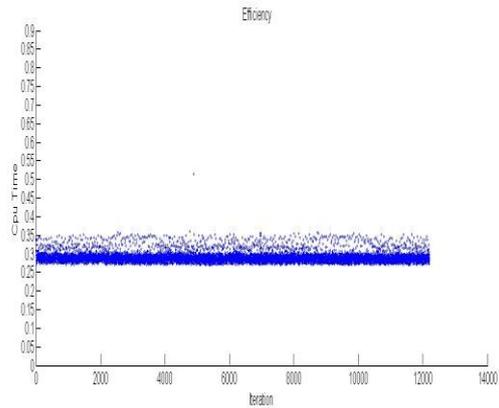

Figure 16: The time efficiency of strong convergence for SA

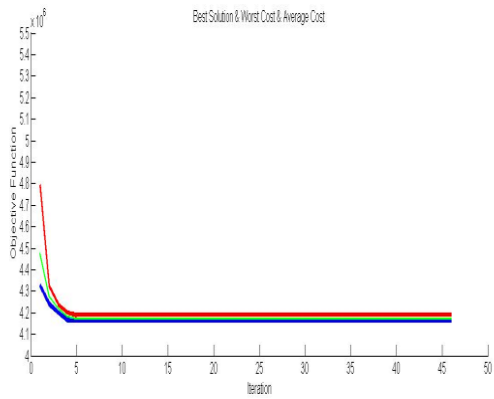 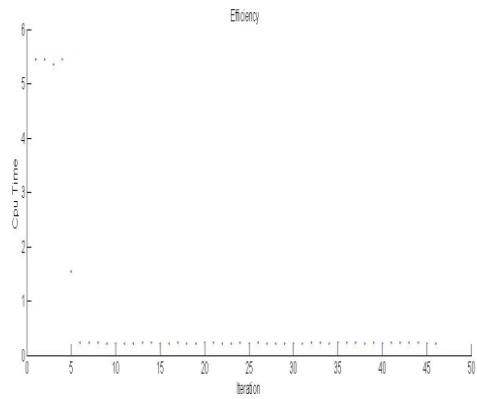

Figure 17: Strong converges rate of Best, average and worst solution for LSH

Figure 18: The time efficiency of strong convergence for LSH

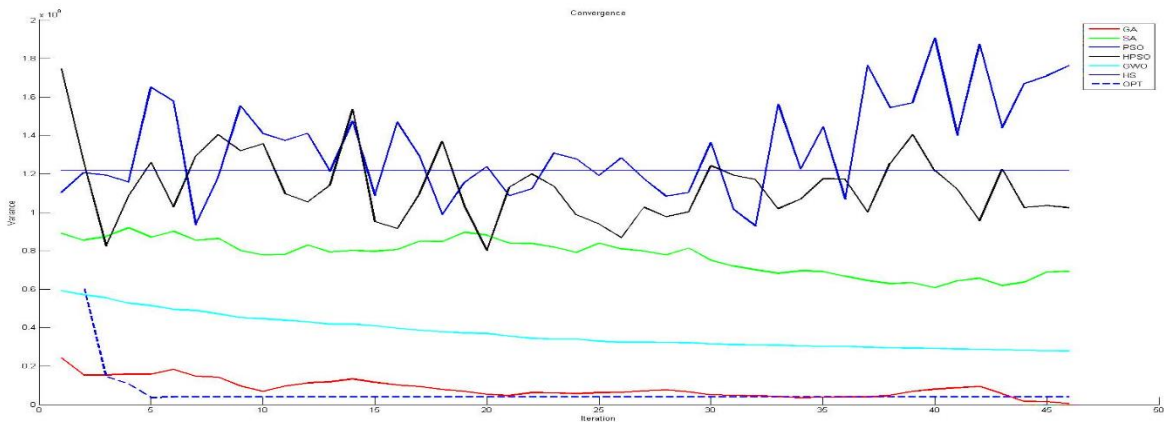

Figure 19: Comparison the trend of strong convergence 50 initial iteration of all algorithms

## Conclusion

In this research, first we implemented seven algorithms on the QAP. Also, we improved the mechanism of three of them (PSO, GWO and 2-Opt). Second, we present a new framework for comparing algorithms. We represent eight factors that show the comparative advantage of each algorithm. By taking into account these factors the test is implemented in 6 test problems. Also, we used a new mechanism for evaluating the strong convergence condition in these algorithms. With respect to this framework we can investigate the advantages of algorithms in order to design a new algorithm that utilize all excellent specifications of these algorithms. Simply, this new algorithm can developed through the hybridization process of these algorithms.

**Biography**

**Reza Tavakkoli-Moghaddam** is a professor of Industrial Engineering in College of Engineering at University of Tehran in Iran. He obtained his Ph.D. in Industrial Engineering from Swinburne University of Technology in Melbourne (1998), his M.Sc. in Industrial Engineering from the University of Melbourne in Melbourne (1994) and his B.Sc. in Industrial Engineering from the Iran University of Science and Technology in Tehran (1989). He was the recipient of the 2009 and 2011 Distinguished Researcher Awards as well as the 2010 and 2014 Distinguished Applied Research Awards by University of Tehran. He was also selected as National Iranian Distinguished Researcher in 2008 and 2010 by the Ministry of Science, Research and Technology (MSRT). Professor Tavakkoli-Moghaddam has published 4 books, 17 book chapters, more than 700 papers in reputable academic journals and conferences.

**Siavash Tabrizian** recieved M.Sc. degree in Industrial Engineering from Department of Industrial Engineering at Sharif University of Technology. He received his B.Sc. degree in Civil Engineering from Department of Civil Engineering at Iran University of Science and Technology. His research interests are combinatorial and computational optimization, and algorithm design.

**Zohreh Raziei** is the M.Sc. student of Industrial Engineering in College of Engineering at University of Tehran in Iran. She received his B.Sc. degree in Applied Mathematics from College of Science, School of Mathematics, Statistics and Computer Science at University of Tehran in Iran. Her research interests include operations research, optimization with applications in disaster management, risk management and healthcare.